\documentclass{article} 

 \usepackage{iclr2023_conference,times}
 
\usepackage{hyperref}
\usepackage{url}

\usepackage[vlined,ruled]{algorithm2e}

\usepackage[utf8]{inputenc} 
\usepackage[T1]{fontenc}    
\usepackage{hyperref}       
\usepackage{url}            
\usepackage{booktabs}       
\usepackage{amsfonts}       
\usepackage{nicefrac}       
\usepackage{microtype}      
\usepackage{thm-restate}

\usepackage{microtype}
\usepackage{graphicx}
\usepackage{subfigure}
\usepackage{changes}

\usepackage{amsmath}
\usepackage{multirow}
\usepackage{hhline}
\usepackage{wrapfig}
\usepackage{floatrow}
\usepackage{caption}
\usepackage{enumitem}
\usepackage[export]{adjustbox}
\usepackage{amsmath,mathtools}

\usepackage[draft]{minted}
\usepackage{pifont}
\definecolor{example_color}{rgb}{0, 0.4, 1}
\definecolor{prompt_color}{rgb}{0.3, 0.3, 0.3}
\definecolor{notation_color}{rgb}{0.8, 0.2, 0.6}
\definecolor{baseline}{rgb}{0.1, 0.4, 0.5}
\definecolor{gray}{rgb}{0.3, 0.3, 0.3}
\definecolor{darkgreen}{rgb}{0.2, 0.6, 0.2}
\definecolor{darkred}{rgb}{1, 0, 0}

\newcommand{\method}{MathPrompter~}
\newcommand{\methodns}{MathPrompter}
\newcommand{\dataset}{MultiArith~}

\newcommand{\correct}{\textcolor{darkgreen}{\ding{51}}}
\newcommand{\wrong}{\textcolor{darkred}{\ding{55}}}
\newcommand\myfontsize{\fontsize{8pt}{10pt}\selectfont}

\usepackage{algorithmic}

\makeatletter
\newcommand{\raisemath}[1]{\mathpalette{\raisem@th{#1}}}
\newcommand{\raisem@th}[3]{\raisebox{#1}{$#2#3$}}
\makeatother

\title{MathPrompter: Mathematical Reasoning using Large Language Models}

\author{Shima Imani, ~~~Liang Du, ~~~Harsh Shrivastava \\
  ~~~~\quad\qquad Microsoft Research, Redmond\\
  \quad Contact: \texttt{shimaimani@microsoft.com}
}
\iclrfinalcopy 
\begin{document}

\maketitle

\begin{abstract}

Large Language Models (LLMs) have limited performance when solving arithmetic reasoning tasks and often provide incorrect answers. Unlike natural language understanding, math problems typically have a single correct answer, making the task of generating accurate solutions more challenging for LLMs. To the best of our knowledge, we are not aware of any LLMs that indicate their level of confidence in their responses which fuels a trust deficit in these models impeding their adoption.
To address this deficiency, we propose `\methodns', a technique that improves performance of LLMs on arithmetic problems along with increased reliance in the predictions. \method uses the Zero-shot chain-of-thought prompting technique to generate multiple Algebraic expressions or Python functions to 
solve the same math problem in different ways and thereby raise the confidence level in the output results.
This is in contrast to other prompt based CoT methods, where there is no check on the validity of the intermediate steps followed. 
Our technique improves over state-of-the-art on the \dataset dataset ($78.7\%\rightarrow92.5\%$) evaluated using 175B parameter GPT-based LLM.
\end{abstract}

\section{Introduction}
\begin{figure*}[t!]
\centering
\includegraphics[height=90mm]{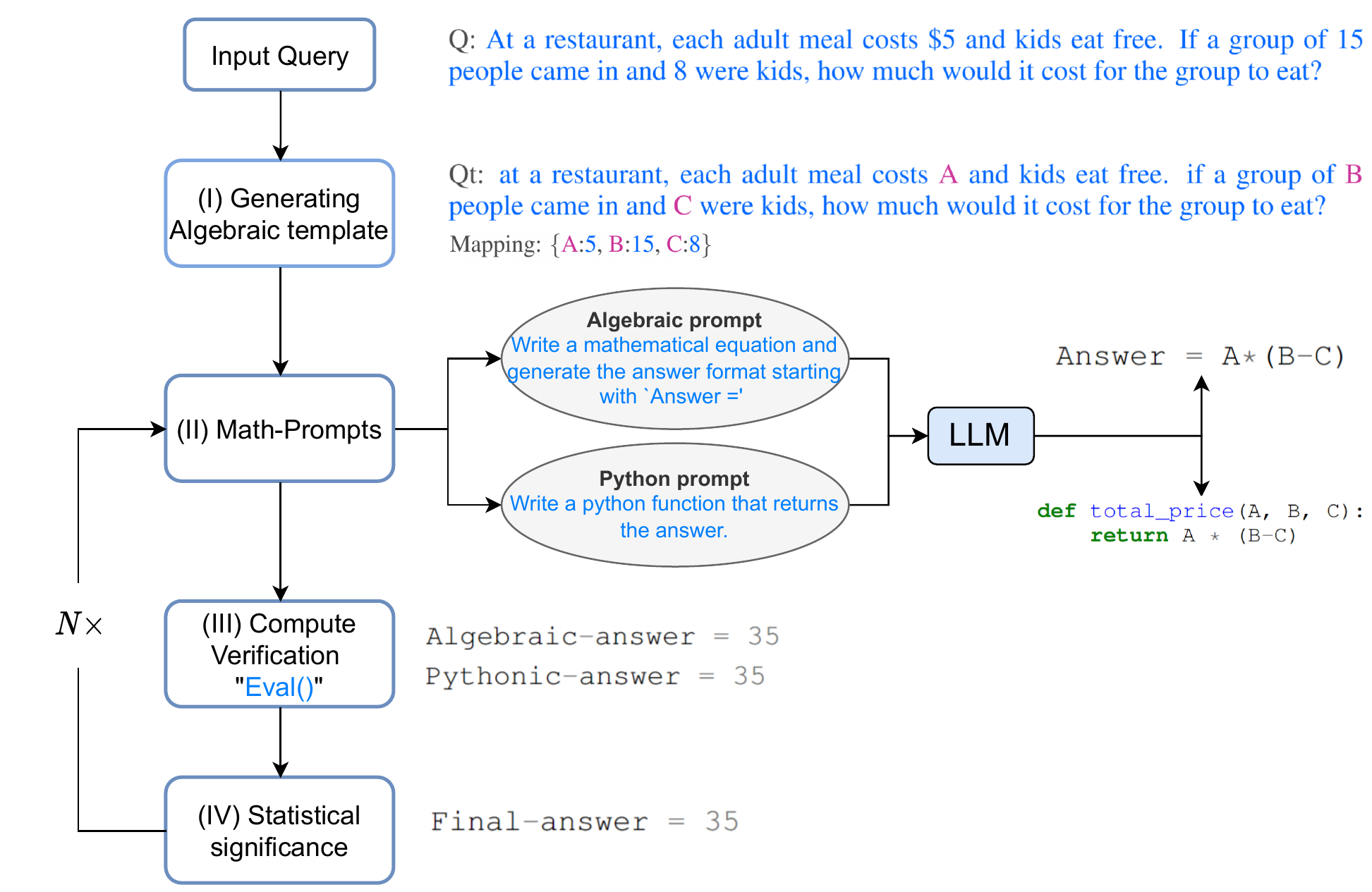}
\caption{\small {\bf\method flow.} We outline the \method process with an example alongside.}~\label{fig:teaser}
\end{figure*}
Recent advancements in natural language processing (NLP) can be attributed to massive scaling of Large Language Models (LLMs)~\cite{vaswani2017attention,devlin2018bert,raffel2020exploring,brown2020language,rae2021scaling,chowdhery2022palm,thoppilan2022lamda}. A very interesting recent discovery that the LLMs are naturally good (in-context) Zero-shot or few-shot learners turned out to be very useful~\cite{brown2020language,liu2021makes,liu2023pre}. This led to the development of `prompting' technique, where the user provides a small context for solving the task at-hand to the LLM. This conditioning of the models on a few examples is termed as few-shot prompting, while providing instructions to solve a task is known as Zero-shot prompting. Extensive research efforts are being poured into designing these prompts, either manually~\cite{schick2020s,reynolds2021prompt} or automatically~\cite{shin2020autoprompt,gao2020making}. Although quite successful for single-step system-I tasks~\cite{stanovich200024,liu2023pre}, the prompting techniques were inadequate in their performance on system-II tasks where multi-step reasoning is required~\cite{rae2021scaling}. As humans, we tend to break down a problem and attempt to solve them step-by-step. Extending this intuition to LLMs led to the development of `chain-of-thought' (CoT) prompting technique~\cite{wei2022chain,wang2022self}. The use of CoT has led to improved performance on a range of NLP tasks~\cite{talmor2018commonsenseqa,gao2020making,patel2021nlp,cobbe2021training,geva2021did,chowdhery2022palm,srivastava2022beyond}



In this work, we investigate Zero-shot-CoT methods for solving mathematical reasoning tasks. To the best of our knowledge, we found the recent work by~\cite{kojima2022large} that proposed a Zero-shot-CoT technique to be the state-of-the-art where they demonstrated a remarkable accuracy improvement on the `MultiArith'~\cite{roy2016solving} data ($17.7\%\rightarrow 78.7\%$). 
Now, we identify two key aspects that lacks in the previous CoT prompting based SOTA, namely (1) Although, the chain-of-thought followed by the model improved the results, but there is \textbf{no check on the validity of the steps followed} by the chain-of-thought prompting and (2) The \textbf{confidence in the predictions} of LLMs are often not provided. In order to address these gap to some extent, we derive inspiration from how we humans solve a math question by breaking it down to a simpler multi-step procedure and make use of multiple ways to validate our approach at each step. Specifically, given a question Q, (I) \textit{Generating Algebraic template}: We first generate its corresponding Algebraic expression Qt that replaces the numerical entries by variables. (II) \textit{Math-prompts}: Then, we provide multiple prompts P to the LLM that can solve Qt analytically in different ways. For eg. P can be `Derive an Algebraic expression' or `Write a Python function' etc. Following this procedure, we end up with P expressions that analytically solves Qt in terms of its variables. (III) \textit{Compute verification}: We then evaluate the P analytical solutions by allotting multiple random values to the Qt variables. (IV) \textit{Statistical significance}: If the solutions of the P analytical functions are in `\textit{consensus}' over $N\sim 5$ different variable choices, then we substitute the original values from Q to obtain the final solution. In the case where there is no definite consensus, we repeat the steps (II), (III) \& (IV). Our method, \methodns, uses 175B parameter LLM called GPT3 DaVinci completion engine \cite{brown2020language}. We were able to improve the accuracy on the \dataset data from $78.7\%\rightarrow 92.5\%$.

\section{Method}




Since the LLMs are generative models, it becomes very tricky to ensure that the generated answers are accurate, especially for mathematical reasoning tasks. We take clues from the process followed by students to solve arithmetic problems. We narrowed down a few steps that students take in order to verify their solutions, namely
\begin{itemize}[wide,labelindent=0pt]
    \item  \textit{Compliance with known results}: By comparing the solution to a known result, one can assess its accuracy and make necessary adjustments. This is particularly useful when the question is a standard problem with a well-established solution. \label{checkanswer}
    
    \item \textit{Multi-verification}: By approaching a problem from multiple perspectives and comparing the results helps to confirm the validity of the solution and ensure that it is both sound and accurate.\label{differentmethods}
    
    \item \textit{Cross-checking}: The process of solving a problem is just as necessary as the final answer. Verifying the correctness of the intermediate steps of the process provide a clear understanding of the thought process behind the solution.\label{intermediatesteps}
        
    \item \textit{Compute verification}: Utilizing a calculator or computer to perform arithmetic calculations can assist in verifying the accuracy of the final answer. \label{calculator}

\end{itemize}

\subsection{\method }


Our proposed method, \methodns, is an attempt to transfer some of this thought process to the LLM answer generation process. Fig.~\ref{fig:teaser} provides a high-level overview of steps followed by \method to solve a mathematical reasoning problem. We use the state-of-the-art GPT-3 DaVinci completion engine \cite{brown2020language} for the question-answering tasks.

We use the following question `{\color{gray} Q}' from the \dataset dataset to demonstrate the problem solving process followed by \methodns.
\begin{quotation}
     \noindent{\color{gray} Q: \textcolor{example_color}{At a restaurant, each adult meal costs \$5 and kids eat free. If a group of 15 people came in and 8 were kids, how much would it cost for the group to eat?}}
\end{quotation}

(I) \textit{Generating Algebraic template}: We begin by transforming the question into its Algebraic form by replacing the numeric entries with variables using a key-value mapping. In this particular instance, the modified question `{\color{gray} Qt}' becomes:

\begin{quotation}
     \noindent{\color{gray} Qt: \textcolor{example_color}{at a restaurant, each adult meal costs {\textcolor{notation_color}{A}} and kids eat free. if a group of {\textcolor{notation_color}{B}} people came in and {\textcolor{notation_color}{C}} were kids, how much would it cost for the group to eat?}}
\end{quotation}
\begin{quotation}   
    \noindent{\color{gray} 
   {Mapping: \{\textcolor{notation_color} {A}:\textcolor{example_color}{5}, \textcolor{notation_color}{B}:\textcolor{example_color}{15}, \textcolor{notation_color}{C}:\textcolor{example_color}{8}}\}
    }
\end{quotation}

(II) \textit{Math-prompts}: We build up on the intuition provided by the multi-verification and cross-checking thought processes mentioned above. We generate analytical solutions of {\color{gray} Qt} using two different approaches, Algebraic way and Pythonic way. We give the following prompts to the LLM to generate additional context for {\color{gray} Qt}



\begin{quotation}
     \noindent{\color{gray} Algebraic prompt: \textcolor{example_color}{
     Write a mathematical equation and generate the answer format starting with `Answer =' }
     }
     
\vspace{3pt} 

     \noindent{\color{gray} Python prompt: \textcolor{example_color}{
     Write a Python function that returns the answer.}
     }
\end{quotation}

The LLM model in response to the above prompts generated the following output expressions
\begin{quotation}
\begin{minted}{Python}
# Algebraic expression output
Answer = A*(B-C)  
\end{minted}
\vspace{3pt}
\begin{minted}{Python}
# Python expression output
def total_price(A, B, C):  
    return A * (B-C)
\end{minted}
\end{quotation}

The above generated analytical solutions gives the user a hint into the `intermediate thought process' of the LLM. Incorporating additional prompts will improve the accuracy and consistency of the results. This will, in turn, enhance the \methodns's ability to generate more precise and effective solutions.

(III) \textit{Compute verification}: We evaluate the expressions generated in the previous step using multiple randomized key-value mappings of the input variables in {\color{gray} Qt}. To evaluate the expressions, we used the Python's \texttt{eval()} method. We compare the outputs to see if we can find a consensus among the answers. This also provides us with a higher level of \textbf{confidence} that the answers are correct and reliable. Once the expressions agree on their outputs, we use the values of the variables in the input {\color{gray} Q} to compute the final answer, as below
\begin{quotation}
    \begin{minted}{Python}
    Algebraic-answer = 35
    Pythonic-answer = 35
    \end{minted}
\end{quotation}


(IV) \textit{Statistical significance}: In order to ensure that consensus is reached among various expressions' output, in our experiments, we repeat the steps (II) \& (III) for $N\sim5$ times and report the most frequent value observed for the answer. 

\section{Experiment}

\begin{table*}[!h]
\centering
\begin{tabular}{l c}
\hline \hline
\bf Model & \makebox[5em]{\bf Accuracy}  \\\hline
\\[\dimexpr-\normalbaselineskip+3pt]
Zero-shot & 17.7 \\
Zero-shot (PaLM 540B) & 25.5 \\
Zero-shot-CoT & 78.7 \\
Zero-shot-CoT (PaLM 540B) & 66.1 \\
Zero-shot-CoT + self consistency (PaLM 540B) & 89.0 \\
Zero-shot-CoT (\textbf{\methodns}) & \textbf{92.5}
\\[\dimexpr-\normalbaselineskip+2pt]\\
\hline
\\[\dimexpr-\normalbaselineskip+3pt]
\textcolor{gray}{Few-Shot (2 samples)} & 33.7 \\
\textcolor{gray}{Few-Shot (8 samples)} & 33.8 \\
\textcolor{gray}{Few-Shot-CoT (2 samples)} & 84.8 \\
\textcolor{gray}{Few-Shot-CoT (4 samples)} & 90.5 \\
\textcolor{gray}{Few-Shot-CoT (8 samples)} & 93.0 \\
\textcolor{gray}{Zero-Plus-Few-Shot-CoT (8 samples)} & 92.8 \\
\\[\dimexpr-\normalbaselineskip+2pt]
\hline
\end{tabular}
\caption {\small \textbf{Accuracy on \dataset dataset}. \method outperforms all the Zero-shot \& Zero-shot-CoT baselines. We emphasize that our model's performance is 
comparable to 540B parameter models as well as the SOTA Few-shot-CoT approaches.
(If not mentioned explicitly, the models in each row consists of 175B parameters. Results are borrowed from~\cite{kojima2022large}. They used Textdavinci-002 (175B) model along with the same 8 examples as described in~\cite{wei2022chain} for Few-shot and Few-shot-CoT settings.)
} \label{tab:baselinePerformance} 
\end{table*}

\subsection{Dataset}
We evaluate \method on \dataset dataset~\cite{roy2016solving}, which is a subset of the Math World Problem Repository \cite{koncel2016mawps}. This dataset is a collection of mathematical problems that are specifically designed to test the ability of machine learning models to perform complex arithmetic operations and reasoning. These problems demand the application of multiple arithmetic operations and logical reasoning to be sucessfully solved.  

\subsection{Baseline}
One of the popular baselines is the standard Zero-shot model by~\cite{brown2020language}. Their train their models in a way that it is able to recognize and classify new objects or classes that it has never seen before during training. This was achieved by utilizing the semantic relationships between classes.

We also compared against the state-of-the-art Zero-shot-CoT prompting model by~\cite{kojima2022large}. This is a very recent approach that addresses the limitations of the standard Zero-shot learning by incorporating a `context of the task' using CoT to improve the performance. Briefly, their method follows this procedure. Given a question Q, the authors use the prompt `Lets think step-by-step' followed by Q to generate a response Z. Then, they use the prompt `The answer (Arabic numericals) is' followed by Z to get their final result.

\begin{table*}[ht]
\centering
\caption{\small Example-based comparison between current state of art Zero-shot-CoT and \methodns.}\label{tab:sample}
\begin{tabular}{|p{0.25\linewidth}|p{0.4\linewidth}|p{0.3\linewidth}|} 
\hline
\textbf{Question} & \textbf{SOTA Zero-shot-CoT} & \textbf{\method} \\\hline
\small{Q: At the fair Adam bought 13 tickets. After riding the ferris wheel he had 4 tickets left. If each ticket cost 9 dollars, how much money did Adam spend riding the ferris wheel?}
& 
\small{A \textcolor{darkred}{(ONE UNNECESSARY STEP ERROR)}: Let’s think step by step. Adam bought 13 tickets. That means he spent 13 * 9 = 117 dollars on tickets. After riding the ferris wheel, he had 4 tickets left. That means he used 9 tickets to ride the ferris wheel. That means he spent 9 * 9 = 81 dollars on riding the ferris wheel. So the answer is 117 - 81 = 36 dollars. Therefore, the answer (arabic numerals) is 36. \wrong }
&{\myfontsize Mapping = {\{\textcolor{notation_color}{A}:\textcolor{example_color}{13}, \textcolor{notation_color}{B}:\textcolor{example_color}{4}, \textcolor{notation_color}{C}:\textcolor{example_color}{9}}\}
\begin{minted}{Python}
# Algabraic answer
Answer = A*(B-C) 
#Python code
def total_price(A, B, C): 
    return A * (B-C)
\end{minted}
}
\small{Answer = 81 \correct}
\\\hline

\small{Q: Each chocolate bar in a box cost \$3. If a box had 9 bars total and Wendy sold all but 3 bars, how much money would she have made?}
& 
\small{A \textcolor{darkred}{(OTHERS; MAXIMUM LIMIT OF TEXT LENGTH)}: Let’s think step by step. Each chocolate bar in a box costs \$3. If a box had 9 bars total, then Wendy sold all but 3 bars. How much money would she have made? We can solve this problem using algebra. Let’s start by creating a variable to represent the number of chocolate bars Wendy sold. We’ll call this variable "x". If Wendy sold "x" chocolate bars, then she would have 9 - x chocolate bars left in the box. We know that each chocolate bar costs \$3, so the total cost of the chocolate bars Wendy sold Therefore, the answer (arabic numerals) is: Wendy would have made \$12.  \wrong}
&
{\myfontsize
Mapping = {\{\textcolor{notation_color}{A}:\textcolor{example_color}{3}, \textcolor{notation_color}{B}:\textcolor{example_color}{9}, \textcolor{notation_color}{C}:\textcolor{example_color}{3}}\}
\vspace{-0.2cm}
\begin{minted}{Python}
# Algabraic answer
Answer = A x (B - C)
#Python code
def money_made(A, B, C):
  return (B-C)*A

\end{minted}
}
\small{Answer = 18 \correct}
\\\hline
\small{Q: Jerry was helping the cafeteria workers pick up lunch trays, but he could only carry 8 trays at a time. If he had to pick up 9 trays from one table and 7 trays from another, how many trips will he make?}
& 
\small{A \textcolor{darkred}{(COMMONSENSE MISTAKE)}: Let’s think step by step. Jerry can carry 8 trays at a time. There are 9 trays on the first table. That means Jerry will need to make 2 trips to the first table. There are 7 trays on the second table. That means Jerry will need to make 1 trip to the second table. In total, Jerry will make 3 trips. Therefore, the answer (arabic numerals) is 3. \wrong}
&
{\myfontsize
Mapping = {\{\textcolor{notation_color}{A}:\textcolor{example_color}{8}, \textcolor{notation_color}{B}:\textcolor{example_color}{9}, \textcolor{notation_color}{C}:\textcolor{example_color}{7}}\}
\begin{minted}{Python}
# Algabraic answer
Answer = (B+C)/A
#Python code
def numTrips(A, B, C): 
  trips = (B + C) // A
  if (B + C) % A != 0:
    trips += 1 
  return trips
\end{minted}
}
\small{Answer = 2 \correct}
\\\hline
\small{Q: Kaleb bought 14 boxes of chocolate candy and gave 5 to his little brother. If each box has 6 pieces inside it, how many pieces did Kaleb still have?}
& 
\small{A: Let’s think step by step. Kaleb bought 14 boxes of chocolate candy. This means that he had 84 pieces of chocolate candy. He gave 5 boxes to his little brother, which means that he gave his little brother 30 pieces of chocolate candy. Kaleb still had 54 pieces of chocolate candy. Therefore, the answer (arabic numerals) is 54 \correct.}
&
{\myfontsize
Mapping = {\{\textcolor{notation_color}{A}:\textcolor{example_color}{14}, \textcolor{notation_color}{B}:\textcolor{example_color}{5}, \textcolor{notation_color}{C}:\textcolor{example_color}{6}}\}
\begin{minted}{Python}
# Algabraic answer
Answer = A*C - B
#Python code
def candy_pieces(A, B, C):
 return A*C - B
\end{minted}
}
\small{Answer = 79 \textcolor{darkred}{(COMMONSENSE
MISTAKE)}\wrong}
\\\hline
\end{tabular}

\end{table*}

\subsection{Results}
\subsubsection{Accuracy comparisons}
Table~\ref{tab:baselinePerformance} compares the performance of the \method against the baseline models. The results of few-shot \& zero-shot learning based approaches are shown. Furthermore, we add the results for models with different number of parameters to get better highlight the significance of our approach.  
Since, \method is a Zero-shot-CoT (175B parameters) method, we choose the state-of-the-art Zero-shot-CoT (175B parameters) model by~\cite{kojima2022large} and a Zero-shot(175B parameters) by~\cite{brown2020language} for fair comparison. We report an accuracy of 92.5\% which is a huge improvement to the other SOTA models with 78.7\% and 17.7\% accuracy, respectively.
\subsubsection{Example comparisons}
Table~\ref{tab:sample} presents a sample set of questions and their respective outputs, intermediate steps, and final answers generated by both \methodns and the current state-of-the-art model \cite{kojima2022large}. For simplicity, only one output of \method for each question is shown for both the Algebraic and Pythonic outputs.

The table highlights areas where \cite{kojima2022large} technique falls short, and where these can be remedied with \method, which was designed to address these issues. For example, the generated answers sometimes have one step of error, which can be avoided by running the model multiple times and reporting the consensus results. Additionally, the reasoning steps in \cite{kojima2022large} can be excessively lengthy, but the Pythonic or Algebraic methods can address this by typically requiring fewer tokens. Furthermore, the reasoning steps may be correct, but the final computation is incorrect. \method address problem by using the Python's \texttt{eval()} method function.

In many cases, the \method generates correct intermediate and final answers. However, there are a few cases, such as the last question in Table \ref{tab:sample}, where both the Algebraic and Pythonic outputs are in agreement, yet erroneous. We plan to address these issues by incorporating additional methods to further enhance the performance of \method.




\section{Conclusions \& Discussions}

We introduced \methodns, a novel approach that improves 
LLM performance on mathematical reasoning problems. It also addresses an important concern of building the user trust to some extent in the LLM predictions. We translated our intuition on how students solve arithmetic problems to a LLM model by utilizing the Zero-shot chain-of-thought prompting technique. \method incorporates ideas like cross-checking the intermediate steps and solving the same math problem using multiple approaches in its design. 
We empirically show that our model is comparable to SOTA Few-shot-CoT models as well as the larger Zero-shot-CoT models that have 540B parameters.
In future, we plan to further evaluate performance on additional datasets and explore incorporating additional prompts into \methodns.
\section{Limitation}
One of the limitations of our work is that while we are running the \method multiple times in different ways to increase the accuracy of our results, this does not always guarantee the correctness of the output. Both Algebraic and Pythonic expressions have the potential to produce the incorrect results, even if the prompt outputs match each other. This is the fail case as shown in the last row of Table~\ref{tab:sample}. Increasing the number of prompts will mitigate this issue. We are currently investigating techniques that can address this issue in a more principled manner.

    

\bibliography{citations}

\bibliographystyle{iclr2023_conference}

\end{document}